\documentclass{article}

\PassOptionsToPackage{numbers, compress}{natbib}

\usepackage[main, final]{neurips_2026}

\usepackage[utf8]{inputenc} 
\usepackage[T1]{fontenc}    
\usepackage{hyperref}       
\usepackage{url}            
\usepackage{amsfonts}       
\usepackage{nicefrac}       
\usepackage{microtype}      

\usepackage{amsmath}
\usepackage{amssymb}
\usepackage{mathtools}
\usepackage{amsthm}
\usepackage{enumitem}
\usepackage{listings}
\usepackage{multirow}
\usepackage[table,xcdraw]{xcolor}
\usepackage[normalem]{ulem}
\usepackage{arydshln}
\usepackage{booktabs}
\usepackage{tcolorbox}
\usepackage{xspace}
\usepackage[utf8]{inputenc}
\usepackage{enumitem}
\usepackage[T1]{fontenc}
\usepackage{wrapfig}

\useunder{\uline}{\ul}{}
\newcommand{\name}[0]{ARTIS\xspace}

\tcbuselibrary{skins, breakable}
\newtcolorbox{PromptBox}[1]{
    colback=gray!5,       
    colframe=gray!75,     
    coltitle=white,       
    fonttitle=\bfseries\sffamily,
    sharp corners,        
    boxrule=0.5pt,        
    title=#1,             
    width=\linewidth,     
    breakable,            
    enhanced,
    before skip=10pt,     
    after skip=10pt       
}

\title{\name: \underline{A}gentic \underline{R}isk-Aware \underline{T}est-Time Scaling via \underline{I}terative \underline{S}imulation}

\author{
  \textbf{Xingshan Zeng}\textsuperscript{1} \quad
  \textbf{Lingzhi Wang}\textsuperscript{2}\thanks{Corresponding Author} \quad
  \textbf{Weiwen Liu}\textsuperscript{3} \quad
  \textbf{Liangyou Li}\textsuperscript{1} \quad
  \textbf{Yasheng Wang} \quad \\
  \textbf{Lifeng Shang}\textsuperscript{1} \quad
  \textbf{Xin Jiang}\textsuperscript{1} \quad
  \textbf{Qun Liu}\textsuperscript{1} \\
  \textsuperscript{1}Huawei Technologies Co., Ltd \\
  \textsuperscript{2}Harbin Institute of Technology, Shenzhen \\
  \textsuperscript{3}Shanghai Jiao Tong University \\
  \texttt{zeng.xingshan@huawei.com} \quad
  \texttt{wanglingzhi@hit.edu.cn} \\
}

\begin{document}

\maketitle

\begin{abstract}
Current test-time scaling (TTS) techniques enhance large language model (LLM) performance by allocating additional computation at inference time, yet they remain insufficient for agentic settings, where actions directly interact with external environments and their effects can be irreversible and costly. We propose \emph{\name}, \emph{\underline{A}gentic \underline{R}isk-Aware \underline{T}est-Time Scaling via \underline{I}terative \underline{S}imulation}, a framework that decouples exploration from commitment by enabling test-time exploration through simulated interactions prior to real-world execution. This design allows extending inference-time computation to improve action-level reliability and robustness without incurring environmental risk. We further show that naive LLM-based simulators struggle to capture rare but high-impact failure modes, substantially limiting their effectiveness for agentic decision making. To address this limitation, we introduce a \emph{risk-aware tool simulator} that emphasizes fidelity on failure-inducing actions via targeted data generation and rebalanced training. Experiments on multi-turn and multi-step agentic benchmarks demonstrate that iterative simulation substantially improves agent reliability, and that risk-aware simulation is essential for consistently realizing these gains across models and tasks.
\end{abstract}

\section{Introduction}
Large language models (LLMs) have recently demonstrated remarkable capabilities in complex reasoning through test-time scaling (TTS), where additional computation at inference time leads to improved answer quality~\cite{zhang2025survey}.
However, many of them are designed for \emph{static, answer-centric} settings, where intermediate reasoning steps incur no real-world risk~\cite{liu20251bllmsurpass405b,luo2024improvemathematicalreasoninglanguage,yao_tree_2023}.
In contrast, an increasing number of applications require LLMs to operate as \emph{agents} that interact with external environments through tool usage~\cite{luo2025large}.
In such agentic settings, model outputs often correspond to irreversible and potentially costly actions. 
In this work, we characterize \emph{risk} as \emph{irreversible, high-impact action failures}, where a single incorrect tool invocation cannot be undone and directly affects downstream execution, potentially leading to cascading failures~\cite{fang-etal-2025-preemptive}.
This exposes a limitation of existing TTS methods, which assume intermediate exploration is free and reversible, overlooking such decision-time risks.

Before taking costly actions, humans often mentally simulate their consequences and refine their plans~\cite{Jensen2024Planning}.
Model predictive control and model-based reinforcement learning follow a similar principle by evaluating hypothetical trajectories before interacting with the environment~\cite{m2023model,morari1999model}.
We argue that agentic models should likewise scale inference-time computation through simulated interaction when actions are costly or irreversible.

Motivated by this observation, prior work has used learned simulators to predict or score candidate action outcomes before commitment~\cite{chae2024web,gu2025webdreamer}.
These methods typically follow a receding-horizon schedule: they commit the selected next action and then replan from the resulting real observation.
While effective for stepwise action selection, this schedule does not explicitly evaluate and aggregate failure evidence across multiple complete task-level attempts.
We introduce \name (\emph{\underline{A}gentic \underline{R}isk-Aware \underline{T}est-Time Scaling via \underline{I}terative \underline{S}imulation}), which scales inference-time computation through \emph{full-trajectory, cross-trial simulation}.
Before any real tool interaction, \name performs multiple complete simulated tool-use attempts, evaluates each attempt at the trajectory level, uses the resulting feedback to refine subsequent trials, and summarizes experience across trials.
Only after this process does the agent begin one committed multi-step execution, allowing downstream dependencies and accumulated failures to be examined before commitment.
Unlike reasoning-centric TTS methods that expand internal reasoning paths, \name allocates additional computation to simulated action trajectories and directly targets action-level correctness.

A key challenge in this framework lies in the reliability of simulation.
Our preliminary analysis shows that naive self-simulation achieved by general LLMs often fail in producing reliable outcomes, leading to degraded performance (Section~\ref{sec:pre-exp}).
In agentic settings, utility is highly asymmetric: most actions have limited impact, but rare failures, particularly irreversible ones, can have disproportionate consequences~\cite{uesatorigorous}.
Simulators optimized for average-case accuracy tend to overlook rare but high-impact failure modes, the dominant source of end-to-end errors in multi-step execution.
Consequently, effective simulation must be \emph{decision-useful} rather than uniformly accurate.
This insight motivates our \emph{risk-aware tool simulator}, which prioritizes fidelity on failure-inducing actions through targeted data generation and rebalanced training. Equipped with this risk-aware simulator, our framework, \name, takes its final form and enables more decision-useful TTS.

We evaluate \name on challenging multi-step agentic benchmarks, including BFCL-v3 Multi-Turn~\cite{berkeley-function-calling-leaderboard} and ACEBench Agent~\cite{chen2025acebench}.
Our results demonstrate that full-trajectory-level iterative simulation substantially improves agent reliability when a high-fidelity simulator is available.
With our risk-aware simulator, agentic TTS consistently outperforms standard inference and conventional TTS baselines, validating the necessity of both iterative simulation and risk-aware simulation.

In summary, this work makes the following contributions:
\begin{itemize}[left=0pt]
\item We introduce \name, a full-trajectory, cross-trial inference framework tailored to action-centric, environment-interactive settings, where actions can carry irreversible, high-impact risks.
\item We develop a risk-aware tool simulator that emphasizes underrepresented failures over average-case accuracy, which enables agents to explore and refine action plans prior to irreversible execution.
\item We show positive aggregate gains on multi-step agentic benchmarks, establishing iterative simulation as an effective form of test-time scaling.
\end{itemize}

\section{Related Work}
\textbf{Agentic Tool Use.}
Tool-augmented LLMs~\cite{liu-etal-apigen-2024,liu-etal-toolace-2025,qin2023toolllm} enable interaction with real-world environments through tool use, forming a critical foundation for autonomous agentic systems. Yet current LLMs continue to exhibit limitations in agentic settings, particularly for complex tasks that require multi-turn tool orchestration and long-horizon decision making~\cite{barres_2-bench_2025,li_tool_2025,berkeley-function-calling-leaderboard}.
A prominent line of research addresses these challenges through large-scale data synthesis, aiming to improve both the diversity and complexity of tool-use behaviors. Existing approaches employ a variety of techniques and pipelines, including multi-agent simulation~\cite{liu-etal-toolace-2025,prabhakar2025apigen,tang_toolalpaca_2023}, graph-based generation~\cite{shim_tooldial_2025,wang-etal-2025-toolflow,yin_magnet_nodate}, leveraging real-world MCP tool ecosystems~\cite{xu_toucan_2025}, and non-sequential generation~\cite{chen_facilitating_2025,zeng2025toolace}.

\textbf{Test-Time Scaling (TTS).}
TTS improves performance by allocating additional computation during inference, which substantially enhances LLM’ reasoning capabilities~\cite{chen2025simpleprovablescalinglaws,wang2024openropensourceframework,wu2024comparativestudyreasoningpatterns}. One line of work adopts training-based TTS, encouraging models to produce more extensive reasoning traces by fine-tuning on data with long chain-of-thought~\cite{chen2025empiricalstudyelicitingimproving,xiang20252reasoningllmslearning} or reflection-augmented examples~\cite{bi2025forestofthoughtscalingtesttimecompute,yu2024improving,zhang2024accessinggpt4levelmathematical}, thereby shifting model behavior from fast responses toward more deliberate reasoning~\cite{guan2025rstar,xi2024enhancingllmreasoningcritique}.
Training-free TTS employs techniques such as tree search, parallel decoding, and sequential revision~\cite{guan2024searchverifyfeedbackgeneration,luo2024improvemathematicalreasoninglanguage,snell_scaling_2024,10.5555/3692070.3694110,wang-etal-2025-chain,wang-etal-2025-stepwise}. These methods typically rely on a verifier to assess intermediate steps, and guide further exploration of the solution space. 
Despite the progress of TTS in reasoning-intensive domains, its extension to agentic systems remains at an early stage. Existing efforts either directly adapt reasoning-centric TTS algorithms to agentic tasks~\cite{chakraborty_role_2025,zeng2025toolacer,zhu2025scaling}, or explore extensions such as test-time context adaptation~\cite{acikgoz2025self,chen2025grounded,zhang2025agentic} and multi-agent scaling~\cite{chen2025tumix,kim2025towards,zhu2025scaling}, largely overlooking the role of environment interaction, which is central in agentic systems.

\textbf{World Modeling.}
World models capture environment dynamics and predict future states, and have been widely studied in multi-modal and embodied settings~\cite{ding2025understanding}. Recent work has explored text-based world models, where LLMs predict structured states, enabling planning and simulation in symbolic environments~\cite{chae2024web,li2025word,wang2024can,wu2025rlvr}. 
The learned world models have also been utilized during inference: WMA~\cite{chae2024web} predicts and scores candidate next-action outcomes, WebDreamer~\cite{gu2025webdreamer} supports multi-step imagined rollouts before stepwise commitment, and DyMo~\cite{guo2025sample} jointly trains tool calling and state prediction for self-verification. In contrast, \name uses a separately reusable simulator for multiple complete tool-use attempts, and evaluate and aggregate trajectory-level experience across trials before one committed execution.

\section{Preliminary Exploration}
\label{sec:pre-exp}
We first explore and address two research questions that directly provide empirical motivation for our design of \name:

\emph{R1: Do agents benefit from performing simulation prior to real execution?}

\emph{R2: Are current general-purpose LLMs capable of effectively serving as tool simulators?}

To answer these two questions, we conduct a preliminary experiment using a naive iterative simulation baseline (not the final choice of our \name framework, and just for fast exploration), where the agent repeatedly performs simulated execution, self-evaluation, and action refinement until the evaluation indicates no further errors.

\begin{wrapfigure}{r}{0.60\textwidth}
\vspace{-0.5cm}
\centering
\includegraphics[width=0.98\linewidth]{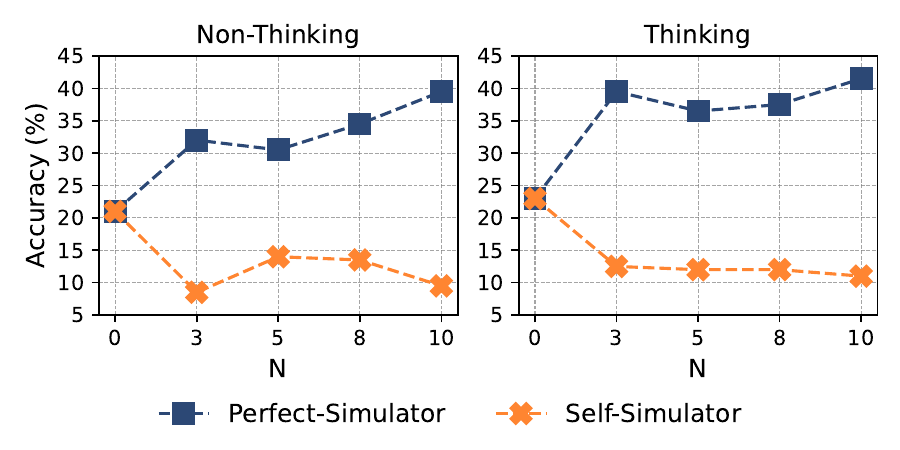}
\vspace{-0.3cm}
\caption{Preliminary results on Qwen3-8B. ``N'' denotes the maximum number of allowed attempts.}
\label{fig:pre-exp}
\vspace{-0.3cm}
\end{wrapfigure}

We consider two simulation settings. In the first, we assume access to a perfect simulator by allowing simulated executions to run directly in the real environment, yielding identical outcomes to actual execution. In the second, the agent itself serves as the simulator, predicting tool responses based on initial environment states and action history through internal reasoning. We instantiate the agent with Qwen3-8B~\cite{yang2025qwen3} and evaluate on the Multi-Turn Base category of BFCL-v3~\cite{berkeley-function-calling-leaderboard}, which uses a Python-based environment that supports multiple attempts. Results are shown in Figure~\ref{fig:pre-exp}.

As shown, with a perfect simulator, performance improves substantially as the maximum number of attempts $N$ increases, providing a positive answer to R1. In contrast, when using the agent itself as the simulator, performance degrades significantly, even falling below the no-simulation baseline ($N=0$), and also remains largely insensitive to $N$. This behavior persists across both non-thinking and thinking modes, indicating that additional reasoning alone does not help. Further analysis of the similarity between ground-truth and simulated outputs (see Table~\ref{tab:sim-compare}) corroborates these findings, collectively answering R2 in the negative.

These observations directly motivate the key design choices in our method: enabling agents to benefit from test-time simulation requires not only iterative refinement, but also a sufficiently accurate and reliable simulator. We therefore next introduce the design of \name.

\section{\name}
We propose \name, a decision-time inference framework that improves the reliability of tool-using agents by simulating multiple complete tool-use trajectories, refining subsequent trials through trajectory-level feedback, and aggregating cross-trial experience before one committed execution.
Unlike conventional TTS methods that primarily expand reasoning paths to obtain a correct final answer, our approach scales computation over simulated interactions to directly target action-level correctness in agentic settings, where tool executions can be irreversible and costly.

\subsection{Problem Setting}

We consider a multi-turn agentic environment where, at each turn, an agent $\mathcal{A}$ is given:
(i) a set of available tools, and (ii) the conversation history including user tasks and interaction history between the agent and the environment.
The agent must fulfill the user request through iterative interaction with the environment via tool calls:
\begin{equation}
a_t \sim \pi_{\mathcal{A}}(a \mid \mathcal{H}_t),
\quad
\mathcal{H}_t = \bigl(\mathcal{H}_0, \{(a_i, o_i)\}_{i=1}^{t-1}\bigr)
\end{equation}
where $\mathcal{H}_0$ denotes the conversation history with the user, $a_i$ is the action (i.e., tool call) produced by the agent in the $i$-th step, and $o_i$ is the corresponding tool response (i.e. the observation).

\subsection{Agentic TTS via Iterative Simulation}
\label{sec:agentic-tts}

\subsubsection{Overall Framework}
\begin{figure}[t]
\centering
\includegraphics[width=0.88\linewidth]{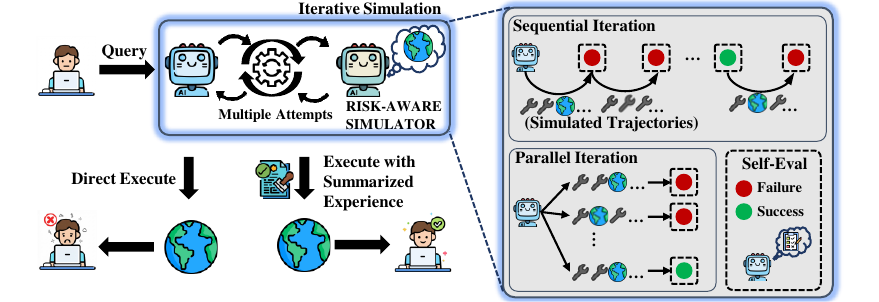}
\vspace{-0.3cm}
\caption{The overall framework for \name, where two modes of Iterative Simulation are shown.}
\label{fig:framework}
\end{figure}

Our idea is straightforward: for each task, the agent performs multiple \emph{simulated attempts} before producing a single \emph{committed execution}.
Each attempt consists of iterative interaction with the simulated environment through multi-step tool calling, followed by self-evaluation and refinement suggestion.
Figure~\ref{fig:framework} illustrates the overall workflow of our method.
Formally, the framework operates in three stages: (i) iterative simulation, (ii) trial summarization, and (iii) final execution.

\subsubsection{Iterative Simulation Loop}

Given the current context (including historical interactions), 
the agent generates up to $N$ candidate attempts to explore potential action space. We consider two modes of iteration, \textit{sequential} and \textit{parallel}, which differ in whether later attempts can observe and leverage the outcomes of earlier ones.

\textbf{Sequential Iteration.}
In the sequential setting, the agent iteratively refines its action proposals by conditioning each new attempt on the outcomes of all previous simulated attempts. This enables adaptive exploration of the action space, where the agent can explicitly avoid previously unsuccessful strategies and progressively improve its decisions. As a result, sequential iteration encourages diverse and non-redundant attempts, leading to more effective coverage of the action space:
\begin{equation}
a_t^{(k)} \sim \pi_{\mathcal{A}}\!\left(
a \mid \mathcal{H}^{(k)}_t,\; \{\mathcal{H}^{(j)}\}_{j=1}^{k-1},\;
\right),
\quad k = 1, \dots, N
\end{equation}
where $\mathcal{H}^{(k)}_t$ is the history before the $t$-th step in the $k$-th attempt, and $\mathcal{H}^{(j)}$ is the entire attempting history for the $j$-th attempt.

However, this advantage comes at the cost of efficiency, as the $N$ attempts must be generated strictly sequentially and the context length grows with the number of attempts. 
When $N$ is large, this can introduce substantial computational overhead and amplify long-context limitations. Therefore, we also propose \textit{Parallel Iteration}.

\textbf{Parallel Iteration.}
In the parallel setting, the agent generates multiple simulated attempts independently, all conditioned on the same conversational context.
Specifically, each attempt is sampled without access to other attempts:
\begin{equation}
a_t^{(k)} \sim \pi_{\mathcal{A}}(a \mid \mathcal{H}^{(k)}_t),
\quad k = 1, \dots, N
\end{equation}
This formulation allows all attempts to be produced simultaneously, enabling efficient parallel computation and avoiding context-length growth.
However, since attempts do not share information, parallel iteration may lead to redundant or highly similar action proposals, particularly when the underlying policy has low sampling diversity.

\textbf{Self-Evaluation.}
A separate evaluation stage assesses the simulated execution, when a simulated attempt is done.
The evaluator determines whether the action is correct and aligned with the task goal, identifies failure causes, and produces a concrete suggestion for improvement.
We leverage the agent itself as the evaluator as most of advanced LLMs are capable of such tasks and even perform comparable to larger models (see results in Figure~\ref{fig:eval-compare}).
The evaluation output consists of a binary correctness signal accompanied by natural-language feedback, which can be incorporated into subsequent refinement steps.

\subsubsection{Summarization of Simulated Attempts}
Before final execution, we must aggregate the outcomes of all simulated attempts. Directly including all results in the prompt can work for strong models (see results in Figure~\ref{fig:ablation}), but is often inefficient and introduces noises that may degrade execution quality. Instead, we introduce a summarization step that distills simulated outcomes into a single natural-language execution strategy $\mathcal{S}$. This summary, also produced by the action agent, aims to maximize robustness while minimizing execution risks. By compressing diverse simulations into a concise guidance signal, it reduces sensitivity to noisy trajectories and supports more risk-averse final decisions.

\subsubsection{Final Execution}

The summarized recommendation is injected into the agent prompt as additional guidance.
The agent then performs one committed multi-step trajectory in the real environment, without launching further simulated trials or rolling back executed actions.
At each step $t$, it selects an action conditioned on the summarized recommendation and the real interaction history:
\begin{equation}
a_t^{\mathrm{real}} \sim \pi_{\mathcal{A}}(a \mid \mathcal{H}_t^{\mathrm{real}}, \mathcal{S}),
\quad
\mathcal{H}_{t+1}^{\mathrm{real}} = \mathcal{H}_t^{\mathrm{real}} \cup \{(a_t^{\mathrm{real}}, o_t^{\mathrm{real}})\}.
\end{equation}
The final committed execution is therefore the complete trajectory
$\tau^{\mathrm{real}} = \{(a_t^{\mathrm{real}}, o_t^{\mathrm{real}})\}_{t=1}^{T}$.

\subsection{Risk-Aware Tool Simulator Training}
\label{sec:simulator}
\begin{wrapfigure}{r}{0.60\textwidth}
\vspace{-0.5cm}
\centering
\includegraphics[width=0.98\linewidth]{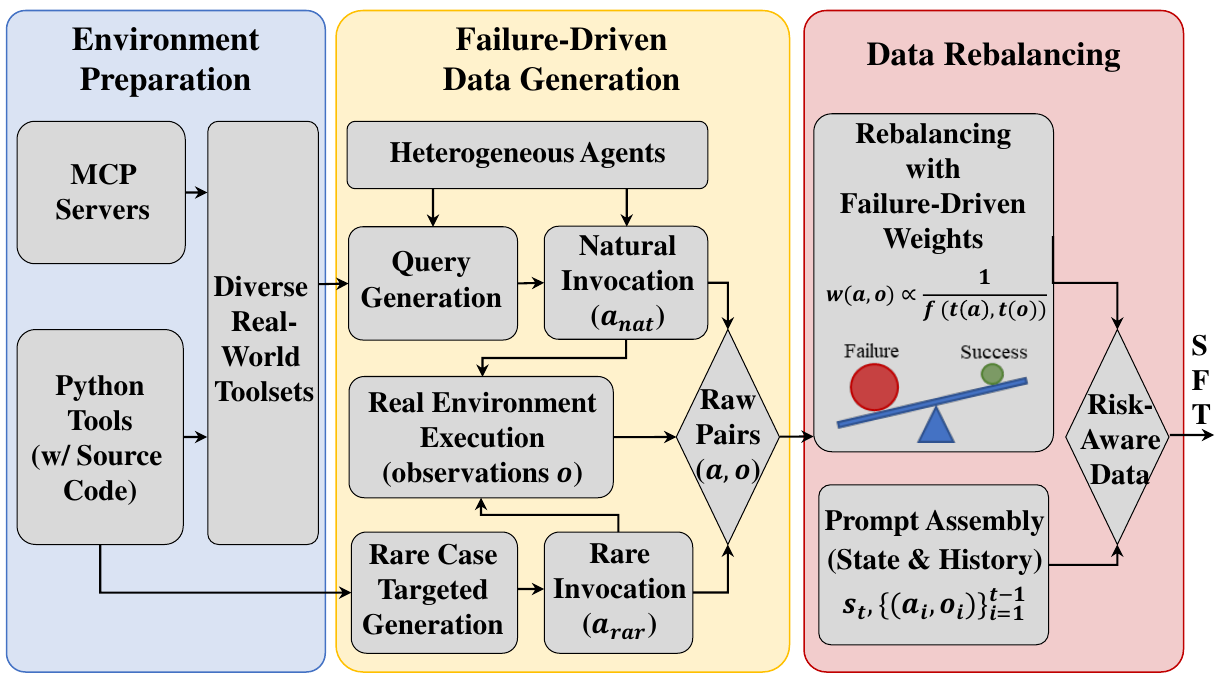}
\vspace{-0.2cm}
\caption{Tool simulator building procedure.}
\label{fig:tool-sim}
\vspace{-0.4cm}
\end{wrapfigure}

The simulator serves as an internal world model that allows agents to assess candidate tool actions before committing to irreversible real-world execution. Rather than optimizing for uniform accuracy, the simulator is explicitly designed to be \emph{decision-useful} by prioritizing the prediction of risky and failure-inducing outcomes. This design is motivated by the asymmetric utility in agentic tool-use scenarios, where rare but high-impact failures dominate overall risk and can mislead agents despite strong average-case performance.
Figure~\ref{fig:tool-sim} displays the overall procedure for building such simulators.

\subsubsection{Simulator Overview}

The simulator is instantiated as an LLM conditioned on a structured prompt representing a hypothetical tool execution.
Formally, the simulator models:
\begin{equation}
p(o \mid a, s),
\end{equation}
where $a$ denotes the tool action, $s$ the environment state, and $o$ the simulated output (including error messages if failed).
The environment state should include what tools are available and any relevant contextual information.

\subsubsection{Failure-Driven Data Generation}

A critical challenge in simulator training is that naive synthetic data generation tends to over-represent successful and trivial cases while under-sampling critical failure cases. To address this, we adopt a \emph{failure-driven data generation} strategy.

\textbf{Environment Preparation.}
To obtain data with real execution outcomes, we first collect a diverse set of real-world tool implementations, primarily from two categories: Python-based functionary tools and MCP servers. For Python-based tools, access to the source code allows us to identify implemented failure modes, which is particularly useful for query generation. Importantly, the source code is used \emph{only for query generation} and is not exposed to the simulator, ensuring a generalized evaluation setting, although it might be useful for simulation (see Appendix~\ref{sec:appendix-simulator-explore}).

\textbf{Failure-Driven Data Generation.}
To obtain natural pairs of tool invocations and their executed outcomes, we first generate user queries conditioned on the collected toolsets, and then prompt tool-using agents to complete these tasks by invoking the corresponding tools. The resulting invocation statements are executed in the real environment to obtain ground-truth outcomes.

To maximize diversity and coverage of failure cases, we employ a heterogeneous set of tool-using agents with varying model sizes and architectures (see Appendix~\ref{sec:appendix-simulator}). This diversity induces a broad range of failure behaviors, particularly from smaller or less capable models.
Nevertheless, certain corner cases remain difficult to trigger through agent execution alone. For these rare errors, we directly prompt LLMs to generate targeted tool invocations, given the relevant implementation code as part of prompt. This allows us to explicitly exercise hard-to-reach failure modes and ensure comprehensive coverage of tool behaviors.

\subsubsection{Simulator Training}
We further rebalance the data to improve coverage across different tools and failure modes. Specifically, we downweight repeatedly observed successful executions and upsample samples corresponding to rare failure cases, by assigning each training instance a sampling weight inversely proportional to the frequency of its outcome type:
\begin{equation}
w(a, o) \propto \frac{1}{\mathrm{freq}(\mathrm{tool}(a), \mathrm{type}(o))},
\end{equation}
where $\mathrm{freq}(\mathrm{tool}(a), \mathrm{type}(o))$ denotes the empirical frequency of the corresponding used tool and outcome type (successful as one single type, failure cases are categorized into different types based on implementation).

Finally, to maintain state consistency when simulating state-based functions, we construct training examples such that each simulated outcome is conditioned on the initial environment state as well as the full execution history:
\begin{equation}
p(o_t \mid a_t, s,  \{(a_i, o_i)\}_{i=1}^{t-1}),
\end{equation}

We then perform supervised fine-tuning (SFT) on the simulator using the rebalanced training samples.

\section{Experiments}
\subsection{Experimental Setup}
\label{sec:exp-setup}
\textbf{Benchmarks.}
We primarily evaluate our method on BFCL-v3~\cite{berkeley-function-calling-leaderboard} and ACEBench~\cite{chen2025acebench}, both of which feature multi-step evaluations and provide isolated execution environments that support multiple iterative attempts. On BFCL-v3, we focus on Multi-Turn categories, while on ACEBench we evaluate the Agent Multi-Step setting.
We also explore effects in single-turn cases for both benchmarks to assess effects across different interaction regimes (Appendix~\ref{sec:appendix-st}).

\textbf{Models.}
Our experiments are mainly conducted using the Qwen3 model family~\cite{yang2025qwen3}, covering model sizes of 4B, 8B, 14B, and 32B parameters. Unless otherwise specified, we adopt the non-thinking mode by default to reduce computational cost, as the performance improvements observed in non-thinking and thinking modes are comparable (see Appendix~\ref{sec:appendix-think-effect}). We further evaluate additional models, including Llama3.1-8B-Instruct~\cite{llama3modelcard}, Llama3.3-70B-Instruct, Qwen2.5-32B-Instruct~\cite{yang2024qwen2} and Mistral-Small-3.2-24B-Instruct-2506\footnote{\url{https://huggingface.co/mistralai/Mistral-Small-3.2-24B-Instruct-2506}}.
For tool simulation, we primarily train the simulator based on Llama3.1-8B-Instruct.

\textbf{Implementation Details.}
To improve exploration of the action space, we set the temperature to 1.0 for the action agent. In contrast, we use a low temperature of 0.01 for the other agents ensuring robustness and consistency. Due to the resource constraints, we set the maximum inference length as 32k. Unless otherwise specified, we set the maximum number of attempts to 5. 
For tool simulator training, please refer to Appendix~\ref{sec:appendix-simulator} for more details. We construct two versions of the training data: one with failure-driven rebalancing and one without, to facilitate controlled comparison. Each training dataset contains approximately 50k instances. 

\begin{table}[t]
\centering
\small
\caption{\label{tab:main_res}
Main comparison results (accuracy, \%) on two benchmarks. Each model is evaluated under two baseline settings (without simulation) and two variants of our method. The best performance in each category within each model is highlighted in \textbf{bold}, while the second best result is {\ul underlined}.}
\setlength\tabcolsep{7.0pt}
\begin{tabular}{lccccccc}
\hline
\multirow{3}{*}{\textbf{Methods}} & \multicolumn{5}{c}{\textbf{BFCL-v3 Multi-Turn}} & \multicolumn{2}{c}{\textbf{ACEBench Agent}} \\
\cmidrule(lr){2-6}\cmidrule(lr){7-8}
 & \multicolumn{1}{c}{\textit{Overall}} &
\multicolumn{1}{c}{\textit{Base}} &
\multicolumn{1}{c}{\begin{tabular}[c]{@{}c@{}}\textit{Miss}\\ \textit{Func}\end{tabular}} &
\multicolumn{1}{c}{\begin{tabular}[c]{@{}c@{}}\textit{Miss}\\ \textit{Param}\end{tabular}} &
\multicolumn{1}{c}{\begin{tabular}[c]{@{}c@{}}\textit{Long}\\ \textit{Context}\end{tabular}} &
\multicolumn{1}{c}{\textit{End-to-End}} &
\multicolumn{1}{c}{\textit{Process}}\\ \hline
\rowcolor[HTML]{EFEFEF}\textit{Qwen3-8B}&12.88& 21.00&11.50&9.50&9.50& 10.00 & 14.60 \\
\hspace{0.5cm} + Weighted BoN & 15.50 & 24.00& {\ul 16.00} & 10.50& 11.50 & 15.00 & 17.60 \\
\hspace{0.5cm} + Sequential Revision & 16.50 & 25.50 & \textbf{16.50} & {\ul 12.50} & 11.50 & 20.00 & 20.00 \\
\hspace{0.5cm} + \name (Parallel) & {\ul 17.75} & {\ul 27.50} &11.50 & \textbf{16.00} & {\ul 16.00} & {\ul 35.00} & {\ul 43.10}\\
\hspace{0.5cm} + \name (Sequential) & \textbf{18.75} & \textbf{29.50} & \textbf{16.50} & {\ul 12.50} & \textbf{16.50} & \textbf{40.00} & \textbf{52.10}\\
\rowcolor[HTML]{EFEFEF}\textit{Qwen3-14B}&18.75& 29.00&18.00&12.00&16.00& 45.00 & 54.20  \\
\hspace{0.5cm} + Weighted BoN & 21.00 & 34.00& {\ul 20.50}& 12.00 & 17.50 & {\ul 55.00} & {\ul 55.80} \\
\hspace{0.5cm} + Sequential Revision & {\ul 24.25} & \textbf{36.50} & 20.00 &{\ul 19.00}&\textbf{21.50} & 50.00 & 50.00\\
\hspace{0.5cm} + \name (Parallel) & 22.50 & {\ul 36.00} & 20.00 & 17.00 & 17.00 & {\ul 55.00} & 55.60\\
\hspace{0.5cm} + \name (Sequential) & \textbf{25.75} & \textbf{36.50} & \textbf{26.50} & \textbf{20.50} & {\ul 19.50} & \textbf{60.00} & \textbf{63.30} \\
\rowcolor[HTML]{EFEFEF}\textit{Qwen3-32B}& 21.75 & 30.50& 23.50& 17.00&16.00& 20.00 & 20.00   \\
\hspace{0.5cm} + Weighted BoN & 24.00 & {\ul 36.00} & 23.00 & 20.00 & 17.00 & 20.00 & 21.30 \\
\hspace{0.5cm} + Sequential Revision & 27.00 & 35.50 & 27.00 & {\ul 22.00} & {\ul 23.50} & 15.00 & 25.00 \\
\hspace{0.5cm} + \name (Parallel) & {\ul 28.75}  & \textbf{41.00} & {\ul 27.50} & 21.00 & \textbf{25.50} & {\ul 25.00} & {\ul 26.30} \\
\hspace{0.5cm} + \name (Sequential) & \textbf{30.50} & \textbf{41.00} & \textbf{34.50} & \textbf{24.00} & 22.50 & \textbf{35.00} & \textbf{36.30} \\
\hline
\end{tabular}

\end{table}

\subsection{Experimental Results}
\subsubsection{Main Results}
Table~\ref{tab:main_res} presents the main results, comparing \name with direct execution and two conventional TTS baselines (implementation details in Appendix~\ref{sec:appendix-baseline}). Since the baselines do not use simulators, TTS is applied at the step level: each step is explored before execution, but no revision is allowed once an action is committed in the real-world environment.

As shown, our sequential iteration variant achieves the best overall performance on both benchmarks across three model scales, demonstrating its effectiveness in complex agentic settings. 
The gains are particularly evident in the more challenging BFCL categories with larger model (Qwen3-32B): it consistently performs best on Miss-Func and Miss-Param, reaching 34.5\% and 24.0\%, respectively. On ACEBench, \name (Sequential) obtains the highest scores in both End-to-End and Process settings for Qwen3-8B and Qwen3-14B, and remains the strongest method overall for Qwen3-32B.

While both baselines outperform direct execution, they remain substantially weaker than our sequential approach. Our parallel variant improves efficiency through parallel decoding at the cost of some performance, yet remains competitive with Sequential Revision and outperforms it in most model-category combinations. 

Given its superior results, all subsequent analyses use the sequential variant unless otherwise specified.

\begin{figure}[t]
    \centering
    \begin{minipage}{0.47\textwidth}
        \centering
        \includegraphics[width=0.95\linewidth]{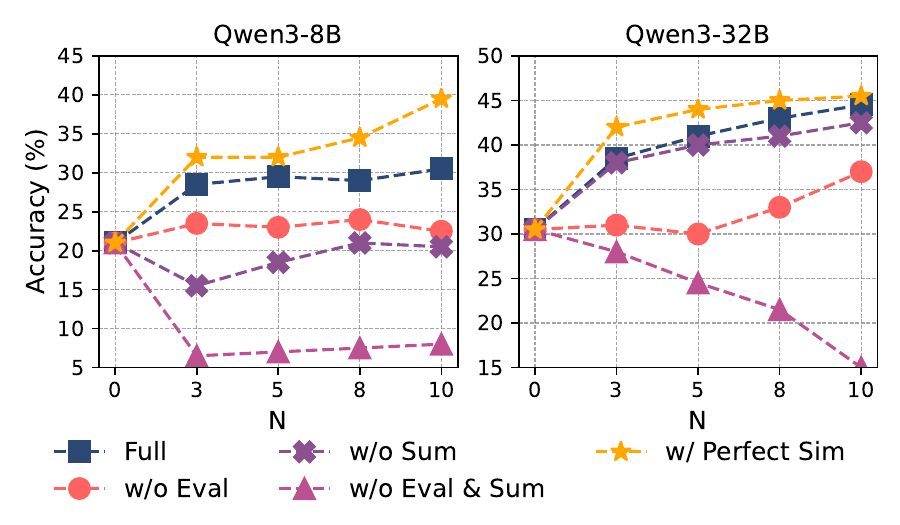}
        \caption{Ablation results on Qwen3-8B and Qwen3-32B, where ``$N$'' denotes the maximum number of allowed attempts and ``Full'' indicates the full \name method.}
        \label{fig:ablation}
    \end{minipage}\hfill
    \begin{minipage}{0.50\textwidth}
        \centering
        \includegraphics[width=0.98\linewidth]{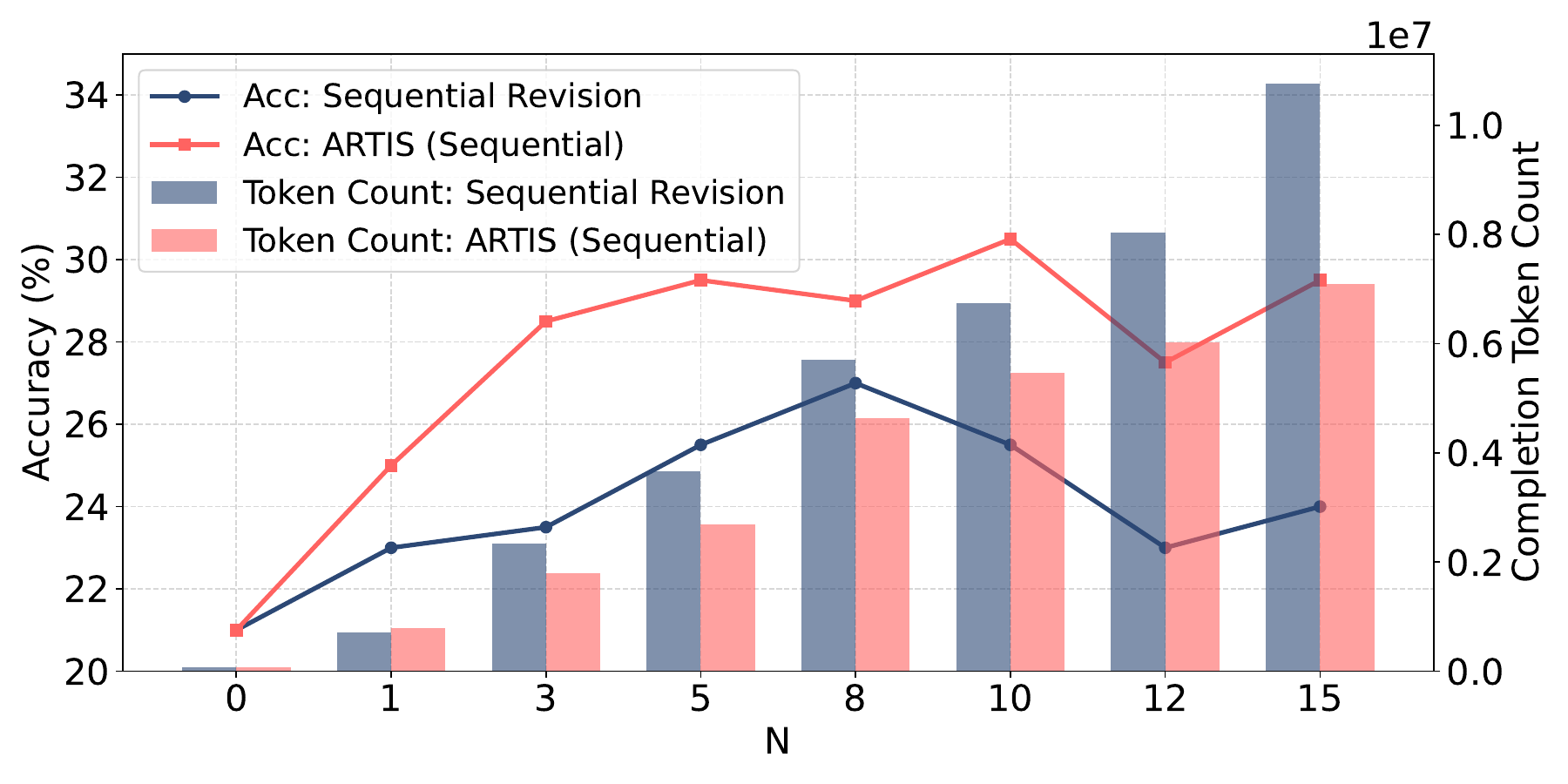}
    \caption{The scaling effects on the accuracy performance and completion token consumption. ``$N$'' denotes the maximum number of allowed attempts.}
    \label{fig:scale-token}
    \end{minipage}
\end{figure}

\subsubsection{Ablation Study}
\label{sec:ablation}
To assess the contribution of each module in \name, we conduct ablations on Qwen3-8B and Qwen3-32B, focusing on the self-evaluation (“Eval”) and summarization (“Sum”) stages. Removing Eval (“w/o Eval”) exposes only raw trajectories to subsequent iterations, while removing Sum (“w/o Sum”) feeds all simulated trajectories directly into the final execution prompt. We also report results with a perfect simulator (“w/ Perfect Sim”). Experiments are conducted in the sequential setting with maximum attempt number $N \in \{0,3,5,8,10\}$ (0 indicates direct execution).

Figure~\ref{fig:ablation} shows that removing either module leads to performance collapse below the baseline. Performance generally improves with larger $N$, although simulation noise still leaves a gap from the perfect simulator. The two models exhibit different sensitivities: Qwen3-8B relies heavily on summarization, without which performance degrades at larger $N$, while Qwen3-32B remains more robust, with the full setup nearly matching the perfect simulator. Notably, the Qwen3-32B “w/o Eval \& Sum” variant declines as $N$ increases, suggesting that excessive raw context without effective evaluation eventually obscures useful signals and harms decision-making.

We also compare our summarization module with alternative designs, further validating the effectiveness of this choice. More details are provided in Appendix~\ref{sec:appendix-sum-choice}.

\subsubsection{Scaling Efficiency}
The ablation study above primarily studies performance scaling across different modules. Here, we further evaluate the scaling efficiency of \name by jointly analyzing performance gains and token consumption as the number of simulation attempts N increases, compared with the corresponding baseline Sequential Revision. Figure~\ref{fig:scale-token} shows results from $N=0$ (direct output) to $N=15$ (statistics are provided in Appendix~\ref{sec:appendix-scale}).

As shown, performance improves with larger
$N$ for both methods, peaking at between 8 and 10. This improvement is accompanied by a substantial increase in token consumption.
Notably, \name not only achieves consistently better performance across all $N$ values, but also demonstrates superior token efficiency compared to the Sequential Revision baseline.
A closer inspection of token usage (see Table~\ref{tab:scale-compare} in Appendix~\ref{sec:appendix-scale}) reveals that the majority of additional tokens originate from self-evaluation, which produces both evaluation outcomes and corresponding natural-language rationales. In \name, self-evaluation is conducted at the task level—evaluating entire action trajectories and their simulated outcomes—resulting in much fewer evaluation calls than the step-level self-evaluation required by Sequential Revision. This design leads to more efficient utilization of inference-time computation while preserving effective performance scaling.

\begin{table}[t]
\centering
\begin{minipage}{0.48\textwidth}
\centering
\caption{\label{tab:sim-compare} Performance, output similarity (\%, vs. a perfect simulator), and high-fidelity ratio (\% of cases with similarity >95\%) for different simulators; all action agents use Qwen3-8B.}
\small
\setlength\tabcolsep{3.6pt}
\begin{tabular}{lccc}
\hline
\multicolumn{1}{l}{\textbf{Simulator}} & \textbf{Perm.} & \textbf{Sim.} & \textbf{HF Ratio} \\ 
\hline
\rowcolor[HTML]{EFEFEF}\textit{w/o Training}&&& \\
Qwen3-8B & 25.5 & 86.0 & 52.6 \\
Qwen3-8B (Thinking) & 21.0 & 88.1 & 58.1 \\
\rowcolor[HTML]{EFEFEF}\textit{Normal Training} &&& \\
Llama3.1-8B-Instruct & {\ul 27.0} & 97.7 & 93.4\\
\rowcolor[HTML]{EFEFEF}\textit{Failure-Driven Training} &&&\\
Llama3.1-8B-Instruct & \textbf{29.5} & {\ul 98.2} & \textbf{95.4} \\
Qwen3-8B &{\ul 27.0} & \textbf{98.3} & {\ul 94.5} \\
\hline
Perfect Simulator & 30.5 & -- & -- \\
\hline
\end{tabular}

\end{minipage}\hfill
\begin{minipage}{0.48\textwidth}
\centering
\caption{\label{tab:diff-backbone} Performance of different backbones on two representative categories of the two benchmarks. ``Mistral3.2-24B-Instruct'' is short for ``Mistral-Small-3.2-24B-Instruct-2506''.}
\small
\setlength\tabcolsep{2.0pt}
\begin{tabular}{lcc}
\hline
\multirow{2}{*}{\textbf{Methods}} & \textbf{BFCL v3} & \textbf{ACEBench} \\ 
& \textit{MT-Base} & \textit{Agent Process} \\ 
\hline
\rowcolor[HTML]{EFEFEF}\textit{Qwen2.5-32B-Instruct} & 25.5 & 15.0 \\
\hspace{0.3cm} + Sequential Revision & 26.5 & 15.7 \\
\hspace{0.3cm} + \name (Sequential) & {\ul 29.0} & {\ul 20.0} \\
\hspace{0.6cm} + Perfect Simulator & \textbf{35.0} & \textbf{22.9}\\
\rowcolor[HTML]{EFEFEF}\textit{Mistral3.2-24B-Instruct} & 7.5 & 11.3 \\
\hspace{0.3cm} + Sequential Revision & 12.5 & 14.1  \\
\hspace{0.3cm} + \name (Sequential) & {\ul 16.0} & {\ul 21.5} \\
\hspace{0.6cm} + Perfect Simulator & \textbf{21.0} & \textbf{23.0} \\
\hline
\end{tabular}
\end{minipage}
\end{table}

\subsubsection{Effectiveness of Trained Simulators}
We compare untrained, normally trained, and failure-driven simulators against a Perfect Simulator, using Qwen3-8B as the action agent on BFCL multi-turn-base. Table~\ref{tab:sim-compare} reports end-to-end performance, cosine similarity to perfect-simulator outputs (using a sentence-transformer\footnote{\url{https://huggingface.co/sentence-transformers/all-MiniLM-L6-v2}}), and the fraction of outputs exceeding 95\% similarity (HF Ratio).
The results show that training substantially improves simulation fidelity over untrained models, while failure-driven training yields the strongest learned simulator and approaches the Perfect Simulator in end-to-end performance (29.5\% v.s. 30.5\%). As for the training base, Llama3.1-8B-Instruct and Qwen3-8B achieve similar output similarity after failure-driven training, with the former attaining a higher HF Ratio and task performance, which then resulting in a higher performance (29.5\% v.s. 27.0\%).

To further evaluate decision-relevant simulator quality, Table~\ref{tab:failure-metrics} reports failure-class macro precision, recall, and F1, together with action accuracy. The table also reports the Risky Action Error Rate (RAER), the fraction of all committed tool calls that are erroneous and state-changing, and the State-changing Action Error Rate (SAER), the fraction of erroneous state-changing calls among all committed state-changing calls. Lower values indicate fewer risky errors. 
The results show that failure-driven training improves failure-class coverage compared with normal training, suggesting that the simulator becomes better at modeling action outcomes. It also achieves the lowest RAER and SAER, indicating that these improvements in simulation quality can translate into safer committed trajectories during actual execution.

\begin{table*}[t]
\centering
\small
\caption{\label{tab:failure-metrics}
Failure-specific simulator evaluation and state-changing action errors on the BFCL-v3(\%). Failure detection metrics compare simulator predictions with real execution outcomes; RAER and SAER evaluate committed trajectories (lower is better).}
\setlength\tabcolsep{7.5pt}
\begin{tabular}{lcccccc}
\hline
\multirow{2}{*}{\textbf{Method}} & \multicolumn{4}{c}{\textbf{Failure Detection}} & \multicolumn{2}{c}{\textbf{State-Changing Errors}} \\
\cmidrule(lr){2-5}\cmidrule(lr){6-7}
& \begin{tabular}[c]{@{}c@{}}\textbf{Action-Level}\\\textbf{Acc.} $\uparrow$\end{tabular}
& \begin{tabular}[c]{@{}c@{}}\textbf{Macro}\\\textbf{Prec.} $\uparrow$\end{tabular}
& \begin{tabular}[c]{@{}c@{}}\textbf{Macro}\\\textbf{Rec.} $\uparrow$\end{tabular}
& \begin{tabular}[c]{@{}c@{}}\textbf{Macro}\\\textbf{F1} $\uparrow$\end{tabular}
& \textbf{RAER} $\downarrow$
& \textbf{SAER} $\downarrow$ \\
\hline
Direct & -- & -- & -- & -- & 39.45  & 71.52  \\
Sequential Revision & -- & -- & -- & -- & 31.91& 59.76  \\
\rowcolor[HTML]{EFEFEF} \name (Normal Training) & {\ul 98.57} & {\ul 81.72} & {\ul 81.12} & {\ul 81.39} & {\ul 24.52} & {\ul 52.25} \\
\rowcolor[HTML]{EFEFEF} \name (Failure-Driven) & \textbf{99.14} & \textbf{95.15} & \textbf{92.58} & \textbf{93.42} & \textbf{23.76} & \textbf{50.99} \\
\hline
\end{tabular}
\end{table*}

\subsubsection{More Analysis}
\label{sec:more-ana}
\textbf{Different Backbones.}
We further investigate the effect of different backbone models to assess the generalizability of our method. Table~\ref{tab:diff-backbone} reports results on Qwen2.5-32B-Instruct and Mistral-Small-3.2-24B-Instruct-2506. Both are relatively large sizes similar to Qwen3-32B. At these scales, both \name and the Sequential Revision baseline achieve substantial performance gains, while \name with a perfect simulator yields the best results across both benchmarks. This indicates that the benefits of agentic test-time scaling are not tied to a specific backbone architecture.

However, as such TTS methods rely on effective self-evaluation and refinement, smaller or less capable models may fail to realize similar gains. As shown in Table~\ref{tab:diff-backbone-more} in the appendix, smaller models such as Qwen3-4B exhibit unstable behavior and may even experience performance degradation. Likewise, earlier models from the Llama3 series (Llama3.1-8B-Instruct and Llama3.3-70B-Instruct) show marginal or negative improvements under both TTS settings, even when provided with a perfect simulator. This indicates that agentic TTS requires a minimum level of reasoning and self-critique capability to be effective, and that insufficient model capacity can limit or negate its benefits.

\begin{wrapfigure}{r}{0.60\textwidth}
\vspace{-0.5cm}
\centering
\includegraphics[width=0.98\linewidth]{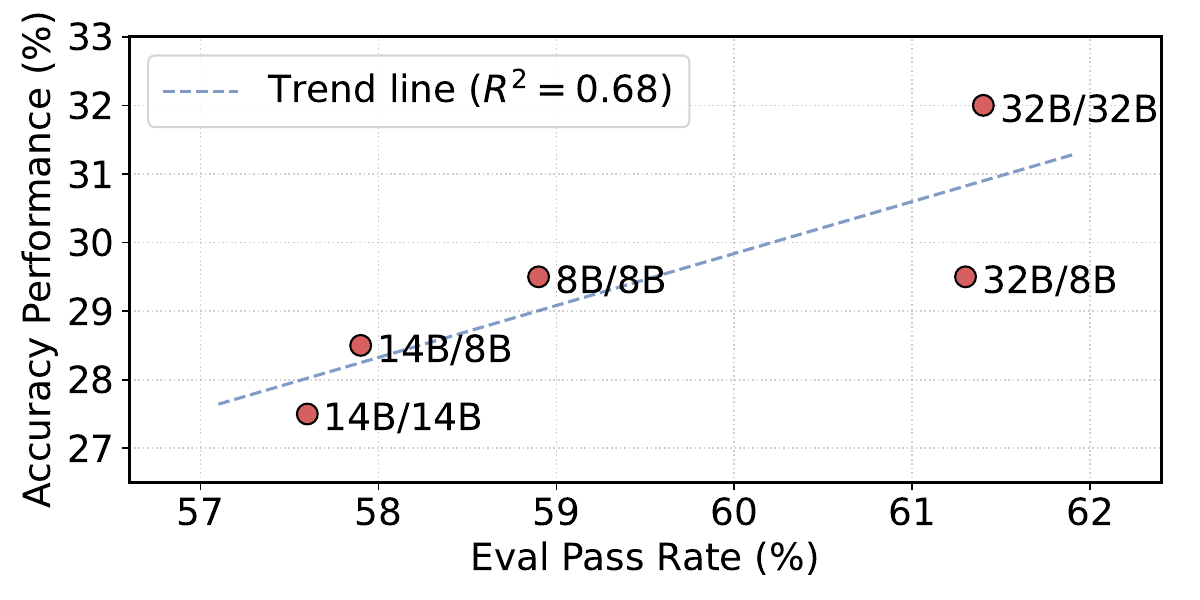}
\vspace{-0.2cm}
\caption{Accuracy and evaluation pass rate comparison for different evaluators and summarizers (all Qwen3 models, noted by ``Evaluator/Summarizer''). Action agents are 8B.}
\label{fig:eval-compare}
\vspace{-0.2cm}
\end{wrapfigure}

\textbf{Evaluator and Summarizer Choices.}
In our default configuration, the action agent itself is also used for self-evaluation and summarization, which constitute two critical components of our framework. This naturally raises a question: whether employing stronger models can further improve performance? To investigate this, we conduct an ablation study in which the action agent is fixed to Qwen3-8B, while the evaluator and summarizer are independently replaced with larger models, namely Qwen3-14B and Qwen3-32B. The corresponding accuracy performance results, as well as the average evaluation pass rate across multiple simulated attempts, are reported in Figure~\ref{fig:eval-compare}.

The results indicate that simply substituting larger models does not consistently lead to improved performance. Notable gains are observed only when both the evaluator and the summarizer are instantiated with Qwen3-32B. Further analysis reveals that performance improvements are closely correlated with the evaluation pass rate (as described by the trend line in Figure~\ref{fig:eval-compare}). In particular, the Qwen3-32B evaluator achieves the highest pass rate, suggesting a stronger ability to identify valid action trajectories. This, in turn, enables more informative and reliable summarization, ultimately benefiting the final execution.

\section{Conclusion}
In this work, we introduce \name, a test-time scaling framework designed for complex agentic scenarios, in which agents are able to iteratively explore multiple complete action attempts within a simulated environment before committing to real-world execution. Central to \name is a risk-aware simulator trained using failure-driven data generation and rebalanced objectives, enabling more accurate modeling of critical error cases. Extensive experiments on two agentic benchmarks demonstrate that \name substantially outperforms existing baselines, generalizes across diverse advanced backbones, and remains effective under different scaling configurations.

\bibliography{example_paper}
\bibliographystyle{plainnat}


\appendix

\newpage

\section{Simulator Training Details}
\label{sec:appendix-simulator}
\textbf{Training Data Details.} 
Our MCP servers primarily source from MCP-Universe~\cite{luo2025mcp}, LiveMCPBench~\cite{mo2025livemcpbench}, and MCPMark~\cite{wu2025mcpmark}, and Python-based function tools from BFCL~\cite{berkeley-function-calling-leaderboard}, $\tau$2-Bench~\cite{barres_2-bench_2025}, API-Bank~\cite{li_api-bank_2023}, and ToolAlpaca~\cite{tang_toolalpaca_2023}. In total, this yields 524 distinct tools. During data generation, we employ Qwen3-32B, Qwen3-8B, Llama3.1-8B-Instruct, Llama3.3-70B-Instruct, and Mistral-Small-3.2-24B-Instruct-2506 as tool-using agents for outcome type coverage. We ensure that each tool is used to generate at least 100 queries. We maintain final training set approximately 50K samples after quality filtering for fair comparison.

\textbf{Training Details.}
Due to resource constraints, we train the simulator using parameter-efficient fine-tuning with LoRA~\cite{hu2022lora}. All model components are LoRA-adapted with rank 16 and scaling factor $\alpha=32$. Training is performed with a global batch size of 64 and a learning rate of $1\times10^{-4}$, following a cosine decay schedule with a 0.1 warmup ratio. Each model is trained for up to three epochs to mitigate overfitting.

\section{Detailed Statistics on Scaling Efficiency}
\label{sec:appendix-scale}
Table~\ref{tab:scale-compare} reports detailed scaling-efficiency statistics, extending Figure~\ref{fig:scale-token} with additional metrics, including API call counts and token counts (both completion and prompt), aggregated overall (``Total'') and broken down by the action agent (``Action'') and the self-evaluation agent (``Self-Eval''). As shown, \name requires substantially fewer self-evaluation calls than Sequential Revision, which needs to perform one self-evaluation per action (when the number of attempts $N$ becomes large, some Sequential Revision runs exceed the 32K context limit, leading to API failures; such cases are directly discarded, resulting in fewer recorded self-evaluation calls than action calls).
This design choice causes Sequential Revision to incur significantly higher API usage and token consumption overall. Nevertheless, because its evaluation operates strictly at the step level and lacks access to task-level planning and long-horizon effects, the resulting performance gains remain limited. In contrast, the improved efficiency and effectiveness of \name highlight the advantages of iterative, task-level simulation for scalable test-time optimization.

\begin{table}[htbp]
    \centering
    \caption{Statistics of the baseline Sequential Revision and \name (Sequential) when scaling $N$.}
    \label{tab:scale-compare}
    \small 
    \resizebox{\textwidth}{!}{ 
    \begin{tabular}{lccccccccccc}
        \toprule
        \multirow{2}{*}{\textbf{Method}} & \multirow{2}{*}{\textbf{N}} & \multirow{2}{*}{\textbf{ACC}} & \multicolumn{3}{c}{\textbf{API Call Count (\#)}} & \multicolumn{3}{c}{\textbf{Completion Token Count (K\#)}} & \multicolumn{3}{c}{\textbf{Prompt Token Count (K\#)}} \\
        \cmidrule(lr){4-6} \cmidrule(lr){7-9} \cmidrule(lr){10-12}
        & & (\%) & Total & Action & Self-Eval & Total & Action & Self-Eval & Total & Action & Self-Eval\\
        \midrule
        Direct & 0 & 21.0 & 2497  & 2497 & -- & 80 & 80 & -- & 15168 & 15168 & -- \\
        \midrule
        & 1 & 23.0 & 5372 & 2686 & 2686 & 718 & 87 & 631 & 48922 &  16471 & 32451 \\
        & 3 & 23.5 & 14588 & 7475 & 7113 & 2339 & 333 & 2005 & 207516 & 87190 & 120326 \\
        Sequential & 5 &25.5 & 26322 & 13245 & 13077 & 3659 & 494 & 3165 & 322449 & 167564 & 154885 \\
        Revision & 8 &27.0 & 39752 & 20176 & 19576 &5697 & 814 & 4883 & 545918 & 287325 & 258592 \\
        & 10 & 25.5 & 51034 & 25871 & 25163 & 6745 & 957 & 5788 & 684898 & 356717 & 328180\\
        & 12 & 23.0 & 61240 & 31124 & 30116 & 8036 & 1181 & 6855 & 797877 & 405014 & 392863\\
        & 15 & 24.0 & 76819 & 39048 & 37771 & 10770 & 1607 & 9163 & 987347 & 506331 & 481015\\
        \midrule
        & 1 & 25.0 & 8124 & 4816 & 745 & 799 & 466 & 171 & 54489 & 29480 & 8759 \\
        & 3 & 28.5 & 17435 & 9425 & 2235 & 1795 & 961 & 503 & 128046 & 63458 & 26328 \\
        \name & 5 & 29.5 & 25809 & 13676 & 3725 & 2689 & 1415 & 830 & 199866 & 100756 & 43880 \\
        (Sequential) & 8 &29.0 & 38470 & 19854 & 5960 & 4643 & 2567 & 1331 & 321349 & 166649 & 70227 \\
        & 10 & 30.5 & 46940 & 24097 & 7450 & 5469 & 3021 & 1658 & 408435 & 217814 & 87796 \\
        & 12 & 27.5 & 55716 &  28471 & 8940 & 6023 & 3076 & 2004 & 506222 & 278984 & 105199 \\
        & 15 & 29.5 & 64357 & 32814 & 11166 & 7094 & 3446 & 2433 & 589988 & 329894 & 130151 \\
        \bottomrule
    \end{tabular}
    }
\end{table}

\section{More Explorations on Simulator Settings}
\label{sec:appendix-simulator-explore}
Section~\ref{sec:pre-exp} examines the limitations of using naive general-purpose LLMs as simulators. Here, we further investigate two potential sources of simulator failure: (1) intrinsic limitations of LLMs, and (2) insufficient environment information. To disentangle these factors, we conduct additional experiments: (1) using a stronger model (Qwen3-32B) as the simulator (noted as “Strong Simulator”), and (2) providing richer inputs, including implementation-level information of the environments, i.e., source code, in the prompt during simulation (noted as “Source Code”). Table~\ref{tab:sim-more-explore} reports the results on the Multi-Turn-Base subset of BFCL-v3.

Both stronger models and richer inputs improve performance, indicating that both factors contribute to simulator quality. However, a noticeable gap remains compared to a perfect simulator, suggesting that accurately modeling rare failure modes is intrinsically challenging. Importantly, for generality, our method does not rely on implementation-level information (e.g., source code), as such information is often unavailable in real-world settings where only tool interfaces are exposed.

\begin{table}[ht]
\small
\centering
\caption{\label{tab:sim-more-explore} Results on general-purpose LLMs as simulators with different settings. }
\setlength\tabcolsep{5.0pt}
\begin{tabular}{lc}
\hline
\multirow{2}{*}{\textbf{Methods}} & \textbf{BFCL v3} \\ 
& \textit{MT-Base} \\ 
\hline
\rowcolor[HTML]{EFEFEF}\textit{Qwen3-8B} & 21.0  \\
\hspace{0.3cm} + Perfect Simulator & 30.5 \\
\hspace{0.3cm} + Self Simulator (Qwen3-8B) & 14.0  \\
\hspace{0.6cm} + Source Code & 22.0  \\
\hspace{0.3cm} + Strong Simulator (Qwen3-32B) & 22.0  \\
\hspace{0.6cm} + Source Code & 26.5  \\
\hline
\end{tabular}
\end{table}

\section{Alternative Choices for Summarization}
\label{sec:appendix-sum-choice}
The summarization module in \name aggregates all simulated trajectories into a high-level textual strategy for guiding real execution, rather than selecting a single trajectory. To further justify this design, we compare summarization with alternative trajectory-selection strategies, including Majority Voting and Best-of-N.
Table~\ref{tab:sum-ablation} presents the results. Although the alternative methods also improve performance, summarization yields more consistent and robust gains overall.

\begin{table}[ht]
\small
\centering
\caption{\label{tab:sum-ablation} Results on comparisons of summarization and other choices. }
\setlength\tabcolsep{5.0pt}
\begin{tabular}{lcc}
\hline
\multirow{2}{*}{\textbf{Methods}} & \textbf{BFCL v3} & \textbf{ACEBench} \\ 
& \textit{MT-Base} & \textit{Agent Process} \\ 
\hline
\rowcolor[HTML]{EFEFEF}\textit{Qwen3-8B} & 21.0  & 14.6 \\
\hspace{0.3cm} + \name (Sequential) & 29.5  & 52.1 \\
\hspace{0.6cm} - Sum & 18.5 & 24.6 \\
\hspace{0.6cm} Replace Sum with Majority Voting & 27.5 & 36.4 \\
\hspace{0.6cm} Replace Sum with Best of N & 29.5 & 28.1 \\
\hline
\end{tabular}
\end{table}

\section{Results on More Different Backbones}

Table~\ref{tab:diff-backbone-more} shows the results with Qwen3-4B, Llama3.1-8B-Instruct, and Llama3.3-70B-Instruct, which serves as a complementary part of Table~\ref{tab:diff-backbone}. As discussed in the main text (Section~\ref{sec:more-ana}), smaller or less capable models struggle to identify reliable improvement signals during TTS, leading to unstable and inconsistent performance across cases.

\begin{table}[ht]
\small
\centering
\caption{\label{tab:diff-backbone-more} Results on smaller or less capable backbones, which may fail to realize observable gains. }
\setlength\tabcolsep{5.0pt}
\begin{tabular}{lcc}
\hline
\multirow{2}{*}{\textbf{Methods}} & \textbf{BFCL v3} & \textbf{ACEBench} \\ 
& \textit{MT-Base} & \textit{Agent Process} \\ 
\hline
\rowcolor[HTML]{EFEFEF}\textit{Qwen3-4B} & 16.5 & 6.7 \\
\hspace{0.3cm} + Sequential Revision & 11.0  & 32.1 \\
\hspace{0.3cm} + \name (Sequential) & 15.0  &  30.6 \\
\hspace{0.6cm} + Perfect Simulator & 24.5 & 26.7 \\
\hline
\rowcolor[HTML]{EFEFEF}\textit{Llama3.1-8B-Instruct} & 11.0 & 6.5 \\
\hspace{0.3cm} + Sequential Revision & 3.5 & 2.5 \\
\hspace{0.3cm} + \name (Sequential) & 6.0 &  5.4 \\
\hspace{0.6cm} + Perfect Simulator &  6.5 & 5.0 \\
\rowcolor[HTML]{EFEFEF}\textit{Llama3.3-70B-Instruct} & 20.0 & 17.5 \\
\hspace{0.3cm} + Sequential Revision & 16.5 & 17.9  \\
\hspace{0.3cm} + \name (Sequential) &  17.5 &  17.5 \\
\hspace{0.6cm} + Perfect Simulator & 21.5 & 14.9 \\
\hline
\end{tabular}
\end{table}

\section{Single-Turn Performance}
\label{sec:appendix-st}
Our main configuration targets multi-step tasks, where iterative simulation can effectively model environment dynamics across successive tool interactions. We further examine the behavior of such TTS methods in single-turn settings, where tasks are completed within a single execution step (possibly involving parallel tool calls). Experiments are conducted on the BFCL~\cite{berkeley-function-calling-leaderboard} Live category and the ACEBench~\cite{chen2025acebench} Single-Turn category. As shown in Table~\ref{tab:single-turn}, both the baseline and our method yield marginal or even negative gains in most cases. This is largely due to the simplicity of single-turn tasks, many of which can already be solved correctly by direct generation without additional inference-time exploration. Moreover, when simulation errors occur, the resulting inaccurate feedback may trigger unnecessary or misleading refinement, leading to over-reflection and degraded final outputs rather than improved decisions.
\begin{table}[ht]
\small
\centering
\caption{\label{tab:single-turn} Accuracy (\%) comparison on single-turn cases of two benchmarks. The best performance in each category is highlighted in \textbf{bold}, while the second best result is {\ul underlined}.}
\setlength\tabcolsep{1.5pt}
\begin{tabular}{lcccccccc}
\hline
\multirow{2}{*}{\textbf{Methods}} & \multicolumn{5}{c}{\textbf{BFCL v3 Live}} & \multicolumn{3}{c}{\textbf{ACEBench Single-Turn}} \\ 
& \textit{Overall}& \textit{Simple}& \textit{Multiple}& \textit{Parallel}& \textit{Para. Multi.} & \textit{Overall}  & \textit{Single Func}  & \textit{Parallel Func} \\
\hline
\rowcolor[HTML]{EFEFEF}\textit{Qwen3-8B} &73.87 &77.52&	73.22&	68.75&	{\ul 66.67}& 62.50 & 73.00 & 52.00\\
\hspace{0.3cm} + Sequential Revision & 74.09 &\textbf{81.40}&	72.27&	\textbf{87.50}&	{\ul 66.67}& 62.50 & 68.00 & 57.00 \\
\hspace{0.3cm} + \name (Sequential) & 74.32 &{\ul 80.62}	&72.65	&{\ul 81.25}&	\textbf{75.00} & 64.00 & 69.00 & 59.00\\
\rowcolor[HTML]{EFEFEF}\textit{Qwen3-32B} &79.57&{\ul 80.62}&	79.68&	{\ul 81.25}&	62.50& {\ul 70.50} & {\ul 75.00} & \textbf{66.00} \\
\hspace{0.3cm} + Sequential Revision & \textbf{80.01} &79.46&	\textbf{80.25}& {\ul 81.25}&	\textbf{75.00}& \textbf{71.50} & \textbf{78.00} & {\ul 65.00}\\
\hspace{0.3cm} + \name (Sequential) & {\ul 79.64}&79.07	&{\ul 79.96}&	\textbf{75.00}	&75.00 & 67.50 & 69.00 & \textbf{66.00} \\
\hline
\end{tabular}
\end{table}

\section{Additional Thinking Effects}
\label{sec:appendix-think-effect}

\begin{figure}[ht]
\centering
\includegraphics[width=0.54\linewidth]{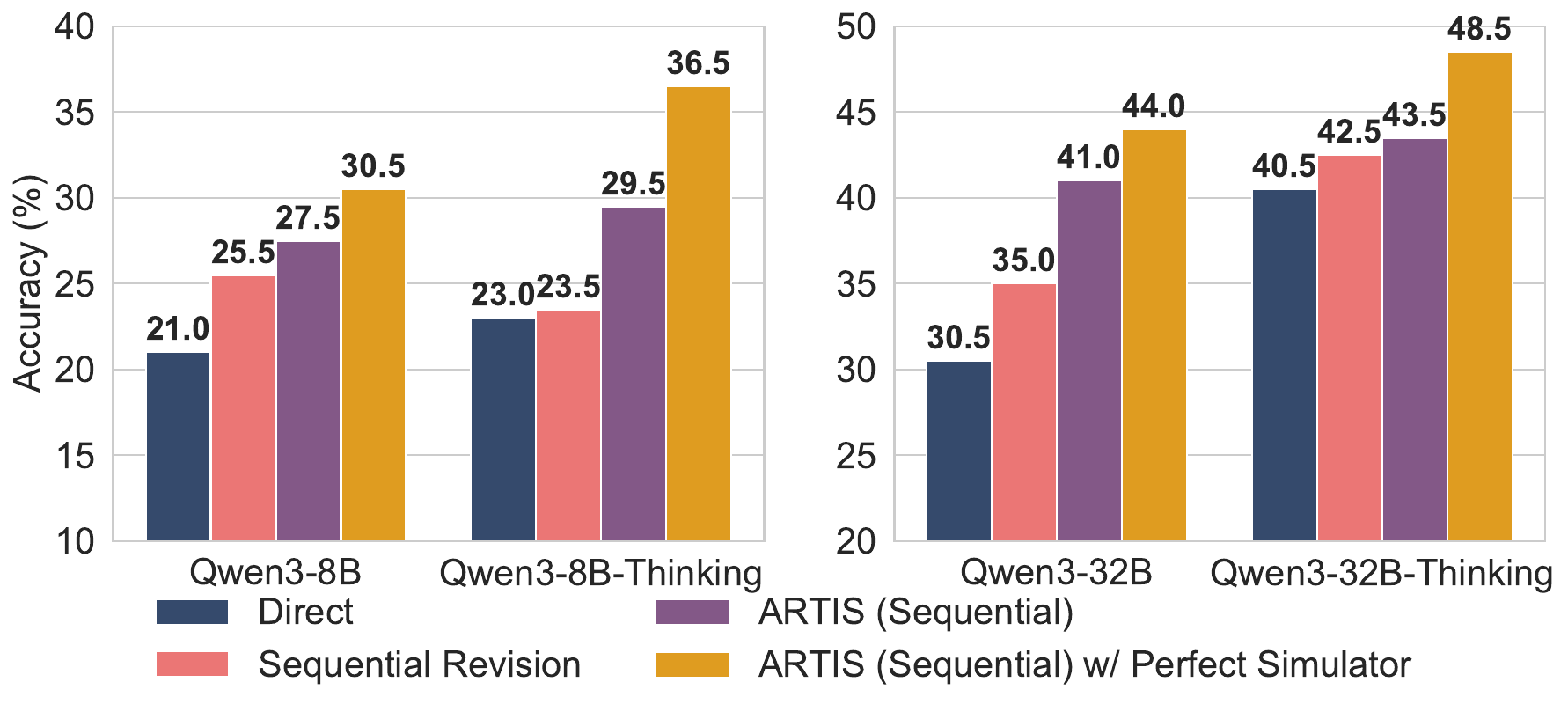}
\caption{Thinking vs Non-Thinking.}
\label{fig:think-or-not}
\end{figure}

We adopt the non-thinking mode for the Qwen3 series models in most experiments for computational efficiency. Here, we further examine the effect of enabling the thinking mode. As shown in Figure~\ref{fig:think-or-not}, additional thinking generally improves overall performance, and TTS methods continue to yield consistent gains under this setting. This demonstrates that our TTS framework is orthogonal to explicit reasoning enhancement: it provides complementary benefits in both thinking and non-thinking modes, rather than relying on extended internal reasoning alone.

\section{Results with Strict Split of Seen/Unseen Tools in BFCL}
\label{sec:appendix-tool-split}
Section~\ref{sec:appendix-simulator} shows that the simulator used in our main experiments is trained on tools that overlap with those evaluated in BFCL, which raises the question of whether the learned simulator can generalize to entirely unseen tools. While ACEBench already provides such an evaluation, as its tools are excluded from simulator training, we further examine this setting within BFCL. Specifically, we retrain the simulator using a strict split over BFCL's 8 tool groups.
\emph{Seen} groups are MessageAPI, TicketAPI, TradingBot, and TravelAPI; \emph{Unseen} groups are GorillaFileSystem, MathAPI, TwitterAPI, and VehicleControlAPI. All tools and generated training examples from the unseen groups are entirely excluded from simulator training, and evaluation cases are partitioned by their tool groups. Table~\ref{tab:seen-unseen-results} compares Direct, Sequential Revision, and \name with normal and failure-driven training, using Qwen3-8B as action agent over three runs.

Failure-driven \name achieves the highest Overall accuracy on both splits, with gains over normal training and both non-simulator baselines. Improvements are most evident in Base and Long Context, although they are not uniform across categories. These aggregate trends are qualitatively consistent with the original evaluation, so we retain the original BFCL results in Table~\ref{tab:main_res}, which do not enforce a strict seen/unseen tool separation, and report the controlled split here as complementary evidence of unseen-tool generalization. ACEBench provides an additional held-out evaluation because none of its tools or environments are included in simulator training.

\begin{table}[h]
\centering
\small
\caption{\label{tab:seen-unseen-results}
Accuracy (\%) on the strict BFCL seen/unseen tool split with Qwen3-8B as the action agent. Results are ``mean $\pm$ std'' over three runs. Best means within each split are in \textbf{bold}.}
\setlength\tabcolsep{4.0pt}
\resizebox{\textwidth}{!}{%
\begin{tabular}{lccccc}
\hline
\textbf{Methods} & \textbf{Overall} & \textbf{Base} & \begin{tabular}[c]{@{}c@{}}\textbf{Miss}\\\textbf{Func}\end{tabular} & \begin{tabular}[c]{@{}c@{}}\textbf{Miss}\\\textbf{Param}\end{tabular} & \begin{tabular}[c]{@{}c@{}}\textbf{Long}\\\textbf{Context}\end{tabular} \\
\hline
\rowcolor[HTML]{EFEFEF}\multicolumn{6}{l}{\textit{Seen Tools}} \\
\rowcolor{white} Direct & 15.31 $\pm$ 0.83 & 27.50 $\pm$ 1.25 & \textbf{14.58 $\pm$ 0.72} & 10.83 $\pm$ 1.44 & 8.33 $\pm$ 0.72 \\
Sequential Revision & 15.94 $\pm$ 0.31 & 28.33 $\pm$ 0.72 & 13.75 $\pm$ 1.25 & \textbf{13.33 $\pm$ 3.82} & 8.33 $\pm$ 0.72 \\
\name (Normal Training) & 15.83 $\pm$ 0.65 & 29.58 $\pm$ 2.60 & 12.08 $\pm$ 1.44 & 10.83 $\pm$ 1.91 & 10.83 $\pm$ 2.60 \\
\name (Failure-Driven) & \textbf{17.92 $\pm$ 0.79} & \textbf{33.33 $\pm$ 1.44} & 12.92 $\pm$ 0.72 & 12.92 $\pm$ 3.82 & \textbf{12.50 $\pm$ 2.17} \\
\hline
\rowcolor[HTML]{EFEFEF}\multicolumn{6}{l}{\textit{Unseen Tools}} \\
\rowcolor{white} Direct & 13.16 $\pm$ 1.83 & 17.54 $\pm$ 4.02 & 11.84 $\pm$ 1.32 & 11.40 $\pm$ 3.04 & 11.84 $\pm$ 0.00 \\
Sequential Revision & 14.25 $\pm$ 0.19 & 18.86 $\pm$ 2.74 & \textbf{13.16 $\pm$ 1.32} & \textbf{12.72 $\pm$ 1.52} & 12.28 $\pm$ 2.01 \\
\name (Normal Training) & 16.23 $\pm$ 1.65 & 22.37 $\pm$ 0.00 & 11.40 $\pm$ 4.23 & 11.40 $\pm$ 3.31 & \textbf{19.74 $\pm$ 2.27} \\
\name (Failure-Driven) & \textbf{17.00 $\pm$ 1.90} & \textbf{22.81 $\pm$ 0.76} & 12.72 $\pm$ 2.01 & \textbf{12.72 $\pm$ 2.01} & \textbf{19.74 $\pm$ 3.48} \\
\hline
\end{tabular}%
}
\end{table}

\section{Matched Simulator-Assisted Baselines}
\label{sec:appendix-simulator-baselines}
To separate the benefit of simulator access from the inference schedule, we compare \name with WMA-style look-ahead~\cite{chae2024web} and SVS-style self-verification inspired by DyMo~\cite{guo2025sample}. All methods share the action model, failure-driven simulator, evaluator backbone, tool information, and unseen-tool split. At each real decision step, WMA-style inference samples candidate actions, simulates their next observations, and scores expected task progress, dependency correctness, and safety. SVS-style inference instead ranks candidates by predicted execution success. Both commit the highest-scoring candidate and replan from the real observation. These are matched inference adaptations rather than reproductions of the original training procedures. For fair comparison, we still use $K=5$: five candidates per real decision step for the baselines and five complete simulated attempts for \name. This matches the exploration count, not total compute. We therefore also report measured inference costs, including LM call times, simulated tool steps, and total tokens.

Table~\ref{tab:simulator-inference-baselines} combines accuracy across all four BFCL categories with per-case compute, including all action-model, simulator, evaluator, and summarizer calls. \name achieves the highest Overall accuracy and leads on three categories while using about 40\% fewer LM calls and 24--26\% fewer input-plus-output tokens. This supports the value of its trajectory-level inference schedule beyond simulator access alone. Gains are modest and category-dependent.

\begin{table}[h]
\centering
\small
\caption{\label{tab:simulator-inference-baselines}
Matched simulator-assisted baseline result comparisons (Accuracy (\%)) on the BFCL-v3 unseen-tool split using Qwen3-8B and the same failure-driven simulator. Compute is reported per case across all four categories. Tokens include input and output (M: million). Best means are in \textbf{bold}.}
\setlength\tabcolsep{5.0pt}
\begin{tabular}{lcccccccc}
\hline
\multirow{2}{*}{\textbf{Method}} & \multicolumn{5}{c}{\textbf{Accuracy} $\uparrow$} & \multicolumn{3}{c}{\textbf{Compute per Case} $\downarrow$} \\
\cmidrule(lr){2-6}\cmidrule(lr){7-9}
& \textbf{Overall} & \textbf{Base} & \begin{tabular}[c]{@{}c@{}}\textbf{Miss}\\\textbf{Func}\end{tabular} & \begin{tabular}[c]{@{}c@{}}\textbf{Miss}\\\textbf{Param}\end{tabular} & \begin{tabular}[c]{@{}c@{}}\textbf{Long}\\\textbf{Context}\end{tabular} & \begin{tabular}[c]{@{}c@{}}\textbf{LM}\\\textbf{Calls}\end{tabular} & \begin{tabular}[c]{@{}c@{}}\textbf{Simulated}\\\textbf{Tool Steps}\end{tabular} & \textbf{Tokens} \\
\hline
WMA-style & 15.57 & 21.05 & \textbf{13.16}  & 12.28  & 15.79  & 196.0 & 83.1 & 1.30M \\
SVS-style & 15.03  & 20.18  & 11.84  & 11.40  & 16.67  & 192.0 & 84.6 & 1.27M \\
\rowcolor[HTML]{EFEFEF}\name & \textbf{17.00} & \textbf{22.81} & 12.72  & \textbf{12.72} & \textbf{19.74} & \textbf{115.7} & \textbf{65.3} & \textbf{0.96M} \\
\hline
\end{tabular}
\end{table}

\section{Limitations}
\label{sec:appendix-limitation}
Our method inherits a fundamental limitation of test-time scaling approaches: improved performance comes at the cost of additional inference-time computation. As a result, \name is less suitable for latency-sensitive scenarios, and is instead better aligned with applications where reducing irreversible or high-impact errors is more important than execution efficiency.

Another limitation is the need for training an additional simulator. In this work, we train the simulator to improve generalization across diverse environments and failure modes. However, as shown in Table~\ref{tab:sim-more-explore}, stronger base models and richer environment information can also provide noticeable gains even without simulator training. This suggests a trade-off between simulator generalizability and training/inference cost. Exploring lightweight and training-free simulation strategies remains an important direction for future work.

\section{Impact Statement}
\label{sec:appendix-impact}
This work addresses the reliability of agentic LLMs that interact with external environments through tool usage, a setting that is increasingly common in automation, software engineering, data management, and decision support systems.
By enabling agents to explore candidate actions through simulation before committing to real execution, the proposed approach may reduce unintended errors and irreversible side effects in real-world deployments.
As agentic systems become more capable and more widely adopted, such improvements in decision-time reliability could contribute to safer and more dependable automation, particularly in environments where failures are costly or difficult to reverse.
At the same time, more effective agentic decision making may accelerate the deployment of autonomous systems, which could have broader societal implications depending on how and where such systems are applied.

From an ethical perspective, our method emphasizes risk-aware decision making by prioritizing the anticipation of underrepresented failure modes.
This design aligns with widely accepted principles of responsible AI.
However, improved simulation and planning capabilities may also increase the effectiveness of autonomous agents in ways that could be misused if deployed without appropriate safeguards.
Our approach does not eliminate the need for external governance mechanisms such as access control, monitoring, and human oversight, especially in safety-critical contexts.
We view iterative simulation as a complementary technical measure that supports, rather than replaces, broader ethical frameworks for the responsible deployment of agentic AI systems.

\section{Prompts}
Here we display the key prompts in our \name framework, including the system prompt for action agent, the user prompt with previous attempting history for action agent, the system and user prompts for self-evaluation stage, the system and user prompts for tool simulator, and the system and user prompt for summarization stage.

\begin{PromptBox}{System Prompt for Action Agent}
You are an expert in composing functions. You are given a question and a set of possible functions. Based on the question, you will need to make one or more function/tool calls to achieve the purpose.

\vspace{5pt}
If none of the functions can be used, point it out. If the given question lacks the parameters required by the function, also point it out.
You should only return the function calls in your response.

\vspace{5pt}
If you decide to invoke any of the function(s), you MUST put it in the format of 

$[\text{func\_name1(params\_name1=value1, params\_name2=value2...), func\_name2(params)}]$

You SHOULD NOT include any other text in the response.

\vspace{5pt}
At each turn, you should try your best to complete the tasks requested by the user within the current turn. Continue to output functions to call until you have fulfilled the user's request to the best of your ability. Once you have no more functions to call, you should output a natural language response, and the system will consider the current turn complete and proceed to the next turn or task.

\vspace{5pt}
Here is a list of functions in JSON format that you can invoke.

\vspace{5pt}
\{functions\}
\end{PromptBox}

\begin{PromptBox}{User Prompt with Previous Attempting History}
\{query\}

\vspace{5pt}
Previous attempts that you can refer to:

\vspace{5pt}
<Attempt>

<Action>\{action1\}</Action>

<Evaluation>\{evaluation1\}</Evaluation>

<Suggestion>\{suggestion1\}</Suggestion>

</Attempt>

\vspace{5pt}
<Attempt>

<Action>\{action2\}</Action>

<Evaluation>\{evaluation2\}</Evaluation>

<Suggestion>\{suggestion2\}</Suggestion>

</Attempt>

\vspace{5pt}
(more attempting history...)
\end{PromptBox}

\begin{PromptBox}{System and User Prompt for Self-Evaluation}
\textbf{SYSTEM:}

\vspace{5pt}
You are an agentic reasoning supervisor. Your job is to evaluate the agent’s last simulated action in a given environment and provide actionable suggestions for improvement.

You must be strict, concise, and goal-oriented. Always assume the agent is allowed to revise its plan.

\vspace{5pt}
\hrule
\vspace{5pt}
\textbf{USER:}

\vspace{5pt}
[Environment Description]

\{tool$\_$documents\}

\vspace{5pt}
[Conversation History]

\{history\}

\vspace{5pt}
[Agent's Simulated Action and Return]

\{simulation\}

\vspace{5pt}
[Your Tasks]
\begin{enumerate}[left=0pt]
    \item Judge whether the agent's action is correct, helpful, and aligned with the user's task goal.
    \item Identify any mistakes or weaknesses in the action.
    \item Explain *why* the action failed or is suboptimal, based strictly on the environment feedback.
    \item Return the exact evaluation result, within the <Result> and </Result> tags, as 1 if correct, else 0.
    \item If not correct, Provide a revised action or an improved action plan for the next simulation try as a suggestion.
\end{enumerate}

\vspace{5pt}
[Output Format]

<Evaluation>

(Detailed judgement and explanation)

</Evaluation>

\vspace{5pt}
<Result>

1/0

</Result>

\vspace{5pt}
<Suggestion>

(Detailed suggestion)

</Suggestion>
\end{PromptBox}

\begin{PromptBox}{System and User Prompt for Tool Simulator}
\textbf{SYSTEM:}

\vspace{5pt}
You are a deterministic environment simulator. Your role is to simulate the tool's response as faithfully as possible, given the initial environment state, history actions and a new action.

You do NOT plan, reason, or suggest improvements.
You only return what the environment would output.

\vspace{5pt}
\hrule
\vspace{5pt}
\textbf{USER:}

\vspace{5pt}
[Environment Description]

\{tool$\_$documents\}

\vspace{5pt}
[Initial Environment State]

\{init$\_$config\}

\vspace{5pt}
[History Actions and Returns]

\{history\}

\vspace{5pt}
[Current Action]

\{action\}

\vspace{5pt}
Simulate the result of executing the action in the environment.

\vspace{5pt}
[Rules]
\begin{itemize}[left=0pt]
    \item Behave strictly like a real environment.
    \item If the action is valid, return the expected output.
    \item If the action is invalid, return a realistic error message.
    \item Do NOT explain your reasoning or suggest any improvements.
    \item Directly output the tool response with a Python list.
\end{itemize}

\vspace{5pt}
[Output Format]

<Output>

[\{json object of tool output 1\}, \{json dict of tool output 2\}, ...]

</Output>
\end{PromptBox}

\begin{PromptBox}{System and User Prompt for Summarization}
\textbf{SYSTEM:}

\vspace{5pt}
You are a senior execution advisor. Your job is to summarize multiple simulated attempts and evaluations into a single, reliable execution recommendation.

You do not simulate tools or evaluate correctness. You produce a final actionable recommendation that will be executed once in a real environment.

\vspace{5pt}
\hrule
\vspace{5pt}
\textbf{USER:}

\vspace{5pt}
[Environment Description]

\{tool$\_$documents\}

\vspace{5pt}
[Conversational History]

\{history\}

\vspace{5pt}
Below are multiple attempts performed in a simulated environment.Each attempt may include:
\begin{itemize}[left=0pt]
\item the action taken
\item the simulated tool return
\item an optional self-evaluation and suggestion
\end{itemize}

[Simulated Attempts and Feedback]

\{Simulation$\_$history\}

\vspace{5pt}
[Your Tasks]
\begin{enumerate}[left=0pt]
    \item Identify the most reliable execution strategy based on all simulated attempts.
    \item Extract key lessons from failures and partial successes.
    \item Determine the safest and most likely-to-succeed action plan.
\end{enumerate}

\vspace{5pt}
[Important Rules]
\begin{itemize}[left=0pt]
    \item Do NOT repeat the full history.
    \item Do NOT output multiple alternative plans.
    \item Produce ONE final recommendation with natural languages.
    \item Be conservative: prefer actions that are robust and verified in simulation.
\end{itemize}

\vspace{5pt}
[Output Format]

\{

\text{ } "recommendation": "...",
    
\text{ } "rationale": "..."
    
\}
\end{PromptBox}

\section{Implementation of Baselines}
\label{sec:appendix-baseline}
We mainly follow \citet{snell_scaling_2024} to implement the two baselines, where we implement and design the prompts directly for agentic tool use.

\textbf{Weighted BoN.}
“Weighted BoN” denotes Weighted Best-of-N, where $N$ candidate solutions are generated in parallel at each step. A scoring agent then evaluates all candidates, and the distinct solution with the highest aggregated score (scores for identical solutions are summed) is selected for real-world execution. After receiving the corresponding tool response, the agent proceeds to the next step using the same procedure until the task is completed. The action agent uses the same system prompt as in our method. Below, we provide the system and user prompts for the scoring agent.

\begin{PromptBox}{Scoring Agent System and User Prompt}
\textbf{SYSTEM:}

\vspace{5pt}
You are a senior execution advisor. Your task is to assess the quality of a simulated response attempt.

You do NOT simulate tools, propose refinements, or generate alternative actions.

You must evaluate how well and suitable the attempt would perform if executed in the real environment.

\vspace{5pt}
\hrule
\vspace{5pt}
\textbf{USER:}

\vspace{5pt}
[Environment Description]

\{tool$\_$documents\}

\vspace{5pt}
[Conversation History]

\{history\}

\vspace{5pt}
[Simulated Attempt]

\{simulation\}

\vspace{5pt}
[Your Tasks]
\begin{enumerate}[left=0pt]
    \item Identify any errors, risks, or inappropriate behaviors in the simulated attempt.
    \item Assign a score from 1 to 10 that reflects the expected quality and reliability of this attempt in a real execution setting (10 = perfect).
    \item Note that the action should only be part of the entire trajectory. You should not assign a low score solely because the goal has not yet been achieved. 
\end{enumerate}

\vspace{5pt}
[Output Format]

\{

\text{ } "evaluation": "(brief explanation)",
    
\text{ } "score": 1 | 2 | 3 | 4 | 5 | 6 | 7 | 8 | 9 | 10
    
\}
\end{PromptBox}

\textbf{Sequential Revision.}
In this baseline, the action agent generates $N$ candidate solutions sequentially, with each generation conditioned on all previously generated solutions. After each candidate is produced, a self-evaluation agent assesses it and provides a score along with qualitative feedback and improvement suggestion. The feedback and suggestion are incorporated into subsequent generations, while the final score is used only for aggregation and not exposed to the action agent. The final action decision is selected as the distinct solution with the highest aggregated score.
For the action agent, we use the same system prompt as in our method, as well as the user prompt with previous attempting history. The self-evaluation agent operates under the system and user prompts described below.

\begin{PromptBox}{Self-Evaluation Agent System and User Prompt}
\textbf{SYSTEM:}

\vspace{5pt}
You are a senior execution advisor. Your task is to assess the quality of a simulated response attempt.

You do NOT simulate tools.

You must evaluate how well and suitable the attempt would perform if executed in the real environment and provide suggestion.

\vspace{5pt}
\hrule
\vspace{5pt}
\textbf{USER:}

\vspace{5pt}
[Environment Description]

\{tool$\_$documents\}

\vspace{5pt}
[Conversation History]

\{history\}

\vspace{5pt}
[Simulated Attempt]

\{simulation\}

\vspace{5pt}
[Your Tasks]
\begin{enumerate}[left=0pt]
    \item Identify any errors, risks, or inappropriate behaviors in the simulated attempt.
    \item Assign a score from 1 to 10 that reflects the expected quality and reliability of this attempt in a real execution setting (10 = perfect).
    \item If not perfect, provide a revised action or an improved action plan for the next simulation try as a suggestion.
    \item Note that the action should only be part of the entire trajectory. You should not assign a low score solely because the goal has not yet been achieved. 
\end{enumerate}

\vspace{5pt}
[Output Format]

\{

\text{ } "evaluation": "(brief explanation)",

\text{ } "suggestion": "(suggestion for improvement)",
    
\text{ } "score": 1 | 2 | 3 | 4 | 5 | 6 | 7 | 8 | 9 | 10
    
\}
\end{PromptBox}


\end{document}